\pdfoutput=1
\documentclass{article}
\usepackage{iclr2026_conference,times}

\usepackage{amsmath,amsfonts,bm}

\def\eqref#1{equation~\ref{#1}}

\def\1{\bm{1}}

\DeclareMathAlphabet{\mathsfit}{\encodingdefault}{\sfdefault}{m}{sl}
\SetMathAlphabet{\mathsfit}{bold}{\encodingdefault}{\sfdefault}{bx}{n}

\DeclareMathOperator*{\argmin}{arg\,min}

 \usepackage{amsthm}
\usepackage{booktabs}
\usepackage{graphicx}
\usepackage{wrapfig}
\usepackage{float}
\usepackage{placeins}
\usepackage{fancyvrb}
\usepackage{flafter}
\usepackage{etoolbox}
\RecustomVerbatimEnvironment{verbatim}{Verbatim}{fontsize=\footnotesize}
\newtheorem{proposition}{Proposition}
\newtheorem{corollary}[proposition]{Corollary}

\usepackage{hyperref}
\hypersetup{hidelinks}
\usepackage{url}

\title{PAST: Privileged Adaptation from Complete Student Trajectories for
On-Policy Self-Distillation}

\author{%
Yangyang Feng$^{1,*}$ \quad Zhuoyan Feng$^{2,*}$ \quad Junlan Chen$^{1}$\\[3pt]
$^1$The Hong Kong University of Science and Technology (Guangzhou)\\
$^2$Sun Yat-sen University\\[3pt]
\texttt{yfeng044@connect.hkust-gz.edu.cn} \quad
\texttt{fengzhy28@mail2.sysu.edu.cn}\\
\texttt{jchen421@connect.hkust-gz.edu.cn}\\[3pt]
$^*$Equal contribution.
}

\iclrfinalcopy
\makeatletter
\patchcmd{\@maketitle}
  {Published as a conference paper at ICLR 2026}
  {Preprint}
  {}
  {\PackageError{past-arxiv}{Unable to replace the ICLR publication header}{}}
\makeatother

\begin{document}

\maketitle

\begin{abstract}
On-policy self-distillation (OPSD) uses a privileged teacher to supervise a reasoning model on prefixes sampled from its own rollouts. Yet each rollout also reveals how the student's response unfolds and whether it succeeds, student-specific hindsight that standard OPSD does not use to form the teacher. We introduce Privileged Adaptation from Student Trajectories (PAST), which treats each completed student trajectory as additional privileged information for the OPSD teacher while leaving the student's distillation prefixes unchanged. PAST preserves the student's next-token distribution on correct trajectories and uses failed trajectories to adapt the teacher toward verified success under student-proximity regularization. We characterize what such a trajectory-conditioned teacher can transfer to a prefix-only student. Forward-KL distillation projects the teacher distributions to their conditional arithmetic mean given the prefix. This projection separates trajectory-specific variation that remains privileged from the mean policy shift available to the student. For correct trajectories, the unclipped population objective also has the frozen student as an ideal distributional fixed point. Across three mathematical reasoning benchmarks, PAST improves the Avg@12 macro average over Vanilla OPSD by 5.6 percentage points. A $2\times2$ factorial study shows gains from both complete-trajectory access and teacher adaptation, while trajectory removal and shuffling confirm that the adapted teacher uses the matching hindsight context.
 \end{abstract}

\section{Introduction}
\label{sec:introduction}

\begin{wrapfigure}{r}{0.52\textwidth}
    \vspace{-0.5\baselineskip}
    \centering
    \includegraphics[width=\linewidth]{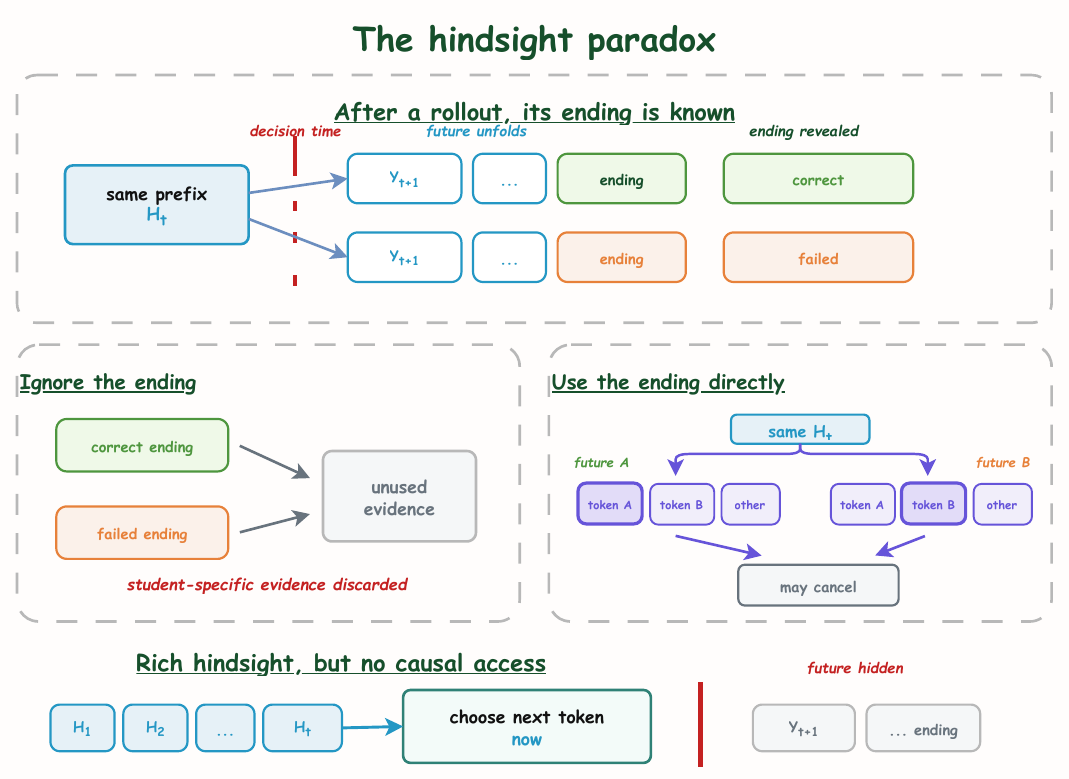}
    \caption{Complete rollouts reveal useful hindsight, but it cannot be copied directly into a prefix-only policy. PAST first adapts the teacher, then distills back to the original prefix.}
    \label{fig:past-intro}
\end{wrapfigure}

On-policy distillation trains a reasoning model on prefixes sampled from its own rollouts, reducing the mismatch between training and inference while retaining token-level teacher supervision \citep{agarwal2024gkd,gu2024minillm}. On-policy self-distillation (OPSD) makes this interface self-contained. A privileged copy of the model teaches its unprivileged policy on those prefixes. Yet, by the time supervision is constructed, the rollout has also revealed the student's complete response and whether it succeeds. Standard OPSD uses task-side privilege to teach on the prefixes, but does not adapt the teacher from how the student's response unfolded \citep{zhao2026opsd}.

Completed responses already support several forms of self-improvement. SD-Zero conditions a reviser on the response and its outcome, H2SD routes successful and failed responses through different hindsight mechanisms, and RSTG supplies teacher guidance to failed zero-variance RL groups \citep{he2026sdzero,cai2026h2sd,han2026distillwherefail}. These methods establish the value of hindsight for revision and guidance. They motivate a broader question. How can complete student trajectories be used more effectively as privileged information in OPSD? For each supervised prefix, the rest of the trajectory is privileged information unavailable to the student.

Using complete trajectories as privilege does not automatically produce useful supervision. Future-dependent targets can conflict at the same prefix and cancel under distillation \citep{zhao2026psopsd,wang2026teachability}. Outcome matters as well. Supervision on correct responses should preserve the student's behavior, whereas supervision on failed responses should move toward verified success without drifting beyond what the student can learn.

We introduce \emph{Privileged Adaptation from Student Trajectories} (PAST), which uses each complete student response as additional privilege for the OPSD teacher. Each cycle freezes the student, samples fresh responses, and adapts one teacher based on final-answer correctness. Full-vocabulary distribution matching preserves the student's next-token distribution on correct responses. On failed responses, verifier-seeking optimization searches for successful continuations while regularization keeps the teacher close to the student. PAST then distills the adapted teacher once on the original student prefixes. Teacher continuations never replace these prefixes, and the next cycle resamples from the updated student.

Our analysis characterizes what a trajectory-conditioned teacher can transfer to a prefix-only student. Forward-KL distillation projects these teacher distributions to their conditional arithmetic mean given the prefix \citep{lin1991divergence,banerjee2005clustering}. This separates privileged variation that the student cannot reproduce from the mean policy shift it can learn, explaining why teacher use of hindsight need not improve the student. On correct trajectories, the unclipped population objective has the frozen student as an ideal distributional fixed point.

Our contributions are summarized below.
\begin{itemize}
    \item We propose PAST, which adapts one privileged teacher from complete correct and failed student trajectories, then distills it onto the student's unchanged on-policy prefixes.
    \item We characterize forward-KL causal projection and separate future-dependent teacher variation from the policy shift available to a prefix-only student.
    \item Across three mathematical reasoning benchmarks, PAST improves the Avg@12 macro average over Vanilla OPSD by 5.6 percentage points. Factorial controls assign positive contributions to trajectory access and teacher adaptation, while trajectory removal and shuffling confirm that the teacher uses the hindsight context.
\end{itemize}

 \section{Preliminaries and Problem Setup}
\label{sec:problem-setup}

PAST combines verifiable policy optimization with on-policy self-distillation. We first define the two training interfaces, then isolate the information change introduced by a complete student trajectory.

\subsection{Verifiable On-Policy Learning}
\label{sec:on-policy-rollouts}

At training cycle $k$, let $(X,Z)\sim\mathcal D$ denote a task and its task-side privilege, such as a reference solution authorized by the training protocol. The frozen student $p_k$ does not observe $Z$ and generates a complete response
\begin{equation}
    Y=(Y_1,\ldots,Y_T)\sim p_k(\cdot\mid X).
    \label{eq:on-policy-rollout}
\end{equation}
A verifier returns the trajectory-level outcome
\begin{equation}
    R=V(X,Y)\in\{0,1\}.
\end{equation}
The verifier may use answer information authorized by the task protocol. It supplies neither token-level labels nor intermediate supervision.

Group Relative Policy Optimization (GRPO) converts such outcomes into an on-policy update \citep{shao2024deepseekmath}. Given $G$ responses $Y^{(g)}$ sampled from an old policy $\mu_{\theta_{\mathrm{old}}}$, let $R_g=V(X,Y^{(g)})$ and
\begin{equation}
    \widehat A_g
    =\frac{R_g-\overline R}{s_R+\epsilon_A},
    \qquad
    \rho_{g,t}(\theta)
    =\frac{\mu_\theta(Y_t^{(g)}\mid X,Y_{<t}^{(g)})}
    {\mu_{\theta_{\mathrm{old}}}(Y_t^{(g)}\mid X,Y_{<t}^{(g)})},
    \label{eq:grpo-advantage-ratio}
\end{equation}
where $\overline R$ and $s_R$ are the group mean and standard deviation. When $s_R=0$, the group-relative advantage is zero. Omitting an optional reference-policy penalty, the loss is
\begin{equation}
\begin{split}
    \mathcal L_{\mathrm{GRPO}}(\theta)
    =-\mathbb E\!\left[
    \frac{1}{G}\sum_{g=1}^{G}\frac{1}{T_g}\sum_{t=1}^{T_g}
    \min\!\left\{
        \rho_{g,t}\widehat A_g,
        \operatorname{clip}(\rho_{g,t},1-\epsilon,1+\epsilon)\widehat A_g
    \right\}
    \right].
    \label{eq:grpo-objective}
\end{split}
\end{equation}
Every token in one response shares the same outcome-derived advantage. PAST uses this group-relative signal only to adapt its teacher on failed student trajectories.

\subsection{On-Policy Distillation}
\label{sec:on-policy-distillation}

On-policy distillation evaluates a teacher on states generated by the student \citep{agarwal2024gkd,gu2024minillm}. At token position $t$, the student-facing state is
\begin{equation}
    H_t=(X,Y_{<t},t).
    \label{eq:student-prefix-state}
\end{equation}
For a teacher distribution $q(\cdot\mid H_t)$ and a candidate student $\pi(\cdot\mid H_t)$, the forward-KL objective is
\begin{equation}
    \mathcal L_{\mathrm{OPD}}(\pi;q)
    =\mathbb E\!\left[
        \frac{1}{T}\sum_{t=1}^{T}
        D_{\mathrm{KL}}\!\left(
            q(\cdot\mid H_t)
            \;\Vert\;
            \pi(\cdot\mid H_t)
        \right)
    \right],
    \label{eq:on-policy-distillation-objective}
\end{equation}
where the expectation follows the student rollout distribution and the teacher is a stop-gradient target. The term \emph{on-policy} refers to the states $H_t$, not to trajectories sampled from the teacher.

\subsection{Privileged On-Policy Self-Distillation}
\label{sec:privileged-opsd}

On-policy self-distillation (OPSD) forms the teacher and student from the same model under different information sets \citep{zhao2026opsd}. Vanilla OPSD supplies the teacher with task-side privilege $Z$ while the student observes only $H_t$.
\begin{equation}
    q_k^{\mathrm{OPSD}}(a\mid H_t,Z),
    \qquad
    p_k(a\mid H_t).
\end{equation}
Its student update instantiates Eq.~(\ref{eq:on-policy-distillation-objective}) with $q=q_k^{\mathrm{OPSD}}$. The sampled response determines the prefixes on which both policies are evaluated, but its unobserved suffix is not part of the Vanilla OPSD teacher context.

\subsection{Trajectory-Privileged Teacher and Causal Student}
\label{sec:privileged-causal-interface}

PAST changes how this privileged teacher is formed. It augments the task-side privilege with the complete student trajectory
\begin{equation}
    U=(Y,Z),
    \label{eq:past-privilege}
\end{equation}
and adapts one teacher $q_\phi(a\mid H_t,U)$. The outcome $R$ selects the adaptation objective but is not included in $U$. Thus the teacher can condition on the suffix of $Y$ beyond $H_t$, while a deployable student $\pi(a\mid H_t)$ remains prefix-only at inference. Appendix~\ref{app:method} compares this interface with the closest on-policy distillation protocols.

After adaptation, PAST freezes the teacher and defines
\begin{equation}
    q_U(a\mid H_t)
    =q_{\phi_{k+1}}(a\mid H_t,U).
\end{equation}
The student is trained on the original rollout states with
\begin{equation}
    \mathcal L_{S,k}(\pi)
    =\mathbb E\!\left[
        \frac{1}{T}\sum_{t=1}^{T}
        D_{\mathrm{KL}}\!\left(
            q_U(\cdot\mid H_t)
            \;\Vert\;
            \pi(\cdot\mid H_t)
        \right)
    \right].
    \label{eq:student-projection-objective}
\end{equation}
Optimizing this objective produces $p_{k+1}$, and the next cycle draws fresh rollouts from the updated student. Section~\ref{sec:method} defines how correct and failed outcomes adapt the teacher. Section~\ref{sec:theory} characterizes the supervision that survives projection to the prefix-only student.

 \section{Method}
\label{sec:method}

\begin{figure}[t]
    \centering
    \includegraphics[width=\linewidth]{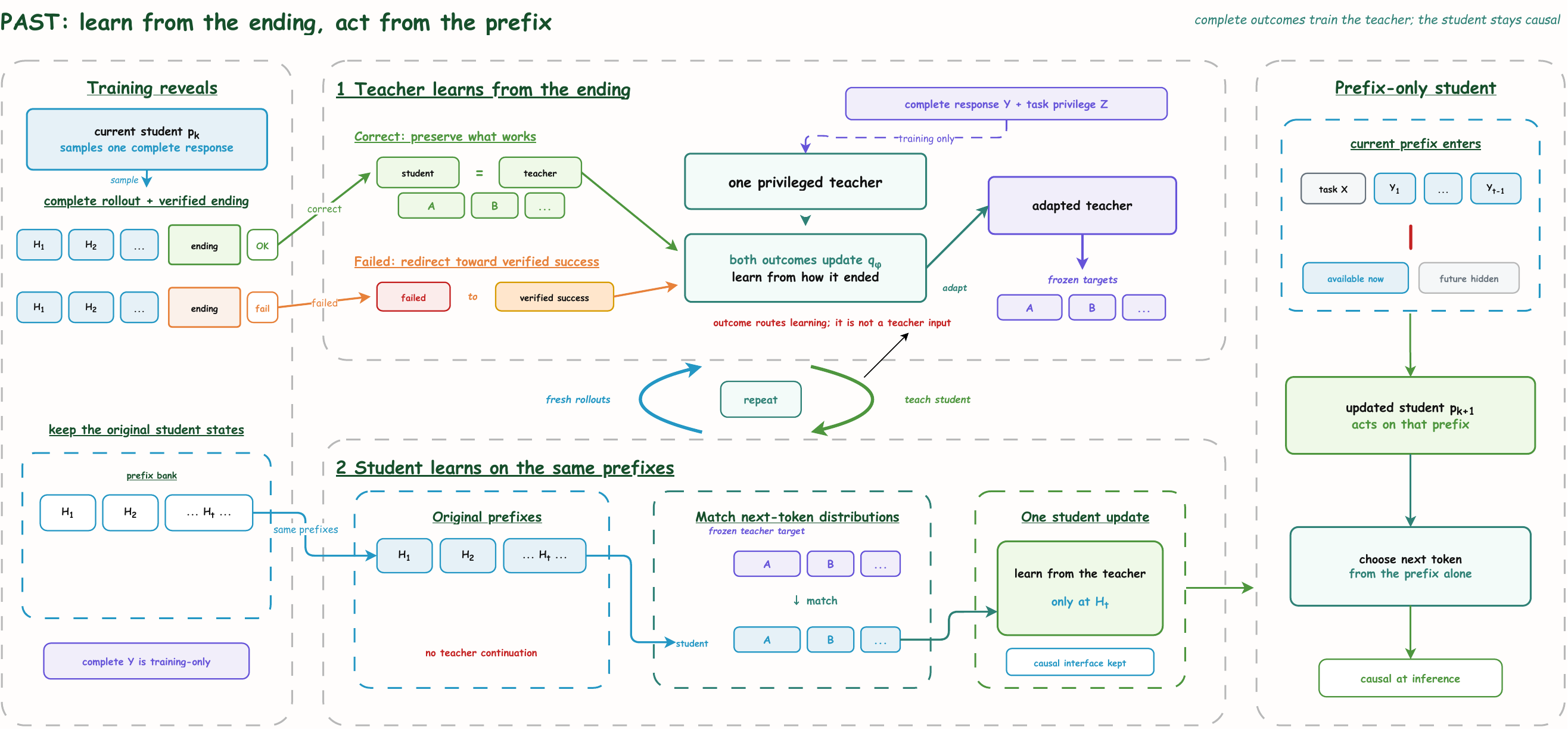}
    \caption{PAST adapts a privileged teacher from complete student responses and outcomes, then distills it on the student's original prefixes. Correct trajectories preserve the current policy. Failed trajectories direct verified improvement under student-proximity regularization, while the student remains prefix-only.}
    \label{fig:past-method}
\end{figure}

PAST turns each fresh student rollout batch into one teacher update followed by one student update. The completed rollout becomes teacher-side privilege, the verifier outcome selects the adaptation objective, and the adapted teacher supervises the student's original prefixes. Correct and failed trajectories train the same teacher. Figure~\ref{fig:past-method} summarizes this cycle.

\begin{figure}[!t]
    \centering
    \includegraphics[width=\linewidth]{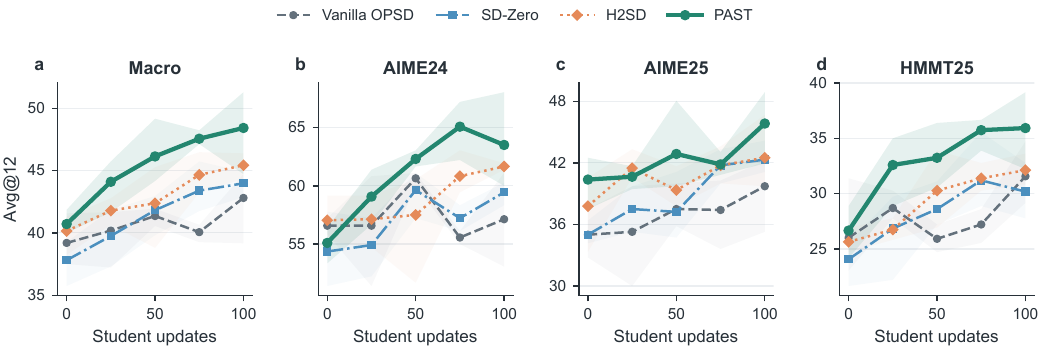}
    \caption{Avg@12 over student updates. Lines show three-seed means, bands show seed ranges, and Macro equally weights the three tasks.}
    \label{fig:main-learning-dynamics}
\end{figure}

\subsection{Complete Student Trajectories as Teacher Privilege}
\label{sec:method-privilege}

At cycle $k$, the frozen student $p_k$ generates $Y\sim p_k(\cdot\mid X)$ without privileged information. PAST gives the teacher the same task-side privilege $Z$ as Vanilla OPSD and adds the complete student response $Y$. The external outcome $R$ selects the training objective but is not included in the teacher input. Both outcome branches therefore adapt one conditional distribution $q_\phi(\cdot\mid H_t,U)$ under the same information interface. We initialize the teacher from the student, $q_0\leftarrow p_0$, so their initial policy difference is zero. Appendix~\ref{app:method} gives the exact input serialization.

\begin{figure}[!t]
    \centering
    \includegraphics[width=\linewidth]{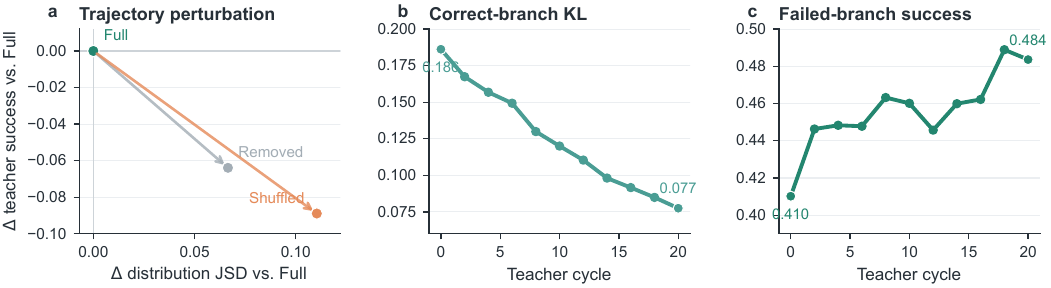}
    \caption{Matching trajectories improve teacher success and compatibility (a). Correct-branch KL falls (b) as failed-branch success rises (c).}
    \label{fig:teacher-mechanism}
\end{figure}

\subsection{Preserving Correct Student Behavior}
\label{sec:method-correct}

A verified-correct trajectory should train the teacher without replacing behavior that already succeeds. PAST therefore matches the frozen student's full next-token distribution on the observed trajectory rather than fitting only its sampled tokens. For distributions $p$ and $q$ at one position, define the OPSD-style clipped contribution
\begin{equation}
    \widetilde D_\tau(p\Vert q)
    =\sum_{v\in\mathcal V}
    \min\!\left\{
        p(v)\log\frac{p(v)}{q(v)},\tau
    \right\}.
    \label{eq:pointwise-clipped-divergence}
\end{equation}
The correct-trajectory loss is
\begin{equation}
    \mathcal L_{+,i}
    =\frac{1}{T_i}\sum_{t=1}^{T_i}
    \widetilde D_{\tau_{\mathrm{OPSD}}}\!\left(
        p_k(\cdot\mid H_{i,t})
        \Vert
        q_\phi(\cdot\mid H_{i,t},U_i)
    \right).
    \label{eq:correct-teacher-loss}
\end{equation}
The threshold and token mask match the corresponding Vanilla OPSD student update. Training uses the clipped quantity for stability and logs the unclipped full-vocabulary KL. No hard-label cross-entropy is added.

\begin{figure}[!t]
    \centering
    \includegraphics[width=\linewidth]{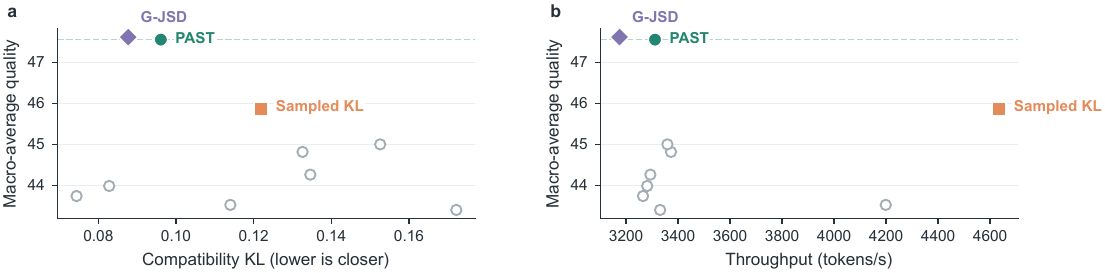}
    \caption{PAST and generalized JSD retain high quality at low compatibility KL, while sampled-token KL trades quality for throughput.}
    \label{fig:objective-tradeoffs}
\end{figure}

\subsection{Adapting on Failed Student Trajectories}
\label{sec:method-failed}

A failed trajectory calls for a success-directed teacher update that remains compatible with the frozen student. The teacher samples complete continuations under $U_i$, and the verifier assigns each continuation a binary reward. A group containing both outcomes supplies the group-relative policy signal $\mathcal L_{\mathrm{GRPO},g}$. In this instantiation of Eq.~(\ref{eq:grpo-objective}), the trainable policy is the trajectory-conditioned teacher $q_\phi$, and the old policy is its frozen snapshot $q_{\phi_k}$ from the start of the teacher update. For each verifier-successful continuation $C_j$ in group $g$, PAST measures its exact proximity to the frozen student
\begin{equation}
    K_j
    =\frac{1}{|C_j|}\sum_s
    D_{\mathrm{KL}}\!\left(
        q_\phi(\cdot\mid G_{j,s},U_i)
        \Vert
        p_k(\cdot\mid G_{j,s})
    \right),
    \label{eq:failed-path-kl}
\end{equation}
where $G_{j,s}$ is the teacher continuation prefix at position $s$. Let $\mathcal S_g$ contain the successful continuations and let $m_g^{\mathrm{mix}}$ indicate a group with both outcomes. An active failed group minimizes
\begin{equation}
    \mathcal L_{-,g}
    =m_g^{\mathrm{mix}}\mathcal L_{\mathrm{GRPO},g}
    +\beta_{\mathrm{KL}}
    \frac{1}{|\mathcal S_g|}\sum_{j\in\mathcal S_g}K_j,
    \qquad \beta_{\mathrm{KL}}=0.05.
    \label{eq:failed-teacher-loss}
\end{equation}
Mixed groups combine success seeking with student proximity. All-success groups retain only the proximity term. Adaptive sampling adds continuations after an initial all-failure draw and skips a trajectory only when a full retry also fails. This rule changes sampling cost rather than the stated population objective. Appendix~\ref{app:method} gives the complete sampler.

PAST first averages examples within each active branch and then averages the active branches
\begin{equation}
    \mathcal L_T
    =\frac{m_+\overline{\mathcal L}_+
    +m_-\overline{\mathcal L}_-}{m_++m_-},
    \label{eq:teacher-active-branch-loss}
\end{equation}
where $m_+$ and $m_-$ indicate whether the corresponding branch contains a valid training term. The observed correct-to-failed ratio therefore does not set the branch weights implicitly.

\subsection{Causal Student Distillation}
\label{sec:method-student}

The student receives the adapted teacher's supervision only on states generated by the frozen student. After one teacher update, PAST freezes $q_{\phi_{k+1}}$ and performs one Vanilla OPSD student update on the original states $H_t$. The implementation applies Eq.~(\ref{eq:pointwise-clipped-divergence}) in the forward direction $q_{\phi_{k+1}}\Vert\pi$ with the baseline vocabulary, threshold, mask, temperature, and reduction. Teacher-generated continuations from the failed branch are not student training paths. The updated policy becomes $p_{k+1}$, cycle-local rollouts are released, and the next cycle samples from the new student. Appendix~\ref{app:method} gives the complete update and snapshot lifecycle.
 \section{Theory}
\label{sec:theory}

PAST gives its teacher access to a complete student trajectory but requires the deployed student to act from its current prefix. We characterize the exact causal target of forward-KL distillation, show why even a perfect privileged teacher can provide no useful student update, and give sufficient conditions under which a teacher change survives projection as a local improvement direction.

\begin{figure}[!t]
    \centering
    \includegraphics[width=0.68\linewidth]{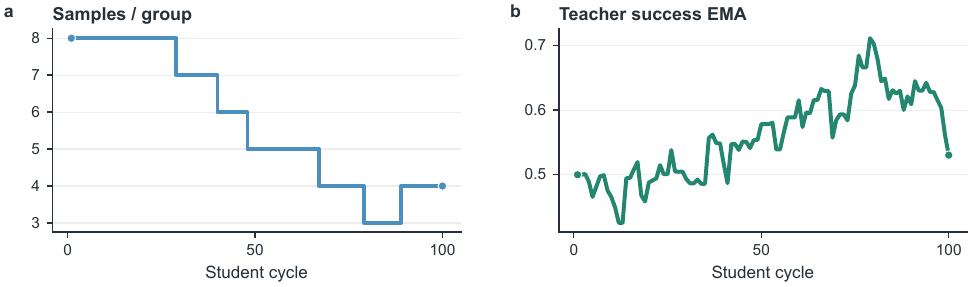}
    \caption{Adaptive sampling reduces its base group size as teacher-success EMA rises over 100 student cycles.}
    \label{fig:adaptive-sampling}
\end{figure}

Fix one training cycle. Let $(H,U)$ follow the distribution induced by the frozen student's on-policy rollouts, where $H$ is a student prefix and $U$ contains the complete trajectory and task-side privilege. Define
\begin{equation}
    q_U(a\mid H)=q_\phi(a\mid H,U),
    \qquad
    \overline q(a\mid H)
    =\mathbb E[q_U(a\mid H)\mid H].
    \label{eq:causal-teacher-barycenter}
\end{equation}
The results use the unclipped population KL under the support and integrability conditions stated in Appendix~\ref{app:theory}.

\begin{proposition}[Privileged-to-causal projection]
\label{prop:causal-projection}
For every prefix-only policy $\pi(a\mid H)$,
\begin{equation}
\begin{split}
    \mathbb E\!\left[D_{\mathrm{KL}}(q_U\Vert\pi)\right]
    ={}&\mathbb E\!\left[D_{\mathrm{KL}}(q_U\Vert\overline q)\right]\\
    &+\mathbb E\!\left[D_{\mathrm{KL}}(\overline q\Vert\pi)\right].
\end{split}
\label{eq:projection-decomposition}
\end{equation}
Consequently, $\overline q$ is the unique minimizer wherever it has positive support.
\end{proposition}

Forward-KL distillation retains the conditional arithmetic mean of the trajectory-conditioned teacher distributions. The first term in Eq.~(\ref{eq:projection-decomposition}) is variation that no prefix-only policy can reproduce. The second is the remaining approximation error to the causal target $\overline q$.

\begin{corollary}[Projection onto a restricted student family]
\label{cor:restricted-causal-projection}
Let $\Pi$ be any class of prefix-only policies for which the displayed expectations are finite. For every $\pi\in\Pi$,
\begin{equation}
\begin{split}
&\mathbb E[D_{\mathrm{KL}}(q_U\Vert\pi)]
-\inf_{\rho\in\Pi}\mathbb E[D_{\mathrm{KL}}(q_U\Vert\rho)]\\
={}&
\mathbb E[D_{\mathrm{KL}}(\overline q\Vert\pi)]
-\inf_{\rho\in\Pi}\mathbb E[D_{\mathrm{KL}}(\overline q\Vert\rho)].
\end{split}
\label{eq:restricted-projection-excess}
\end{equation}
The two objectives have the same minimizing sequences and, whenever a minimum is attained, the same minimizers over $\Pi$. Restricting the student family changes how closely it can represent $\overline q$, but not the causal target selected by forward-KL distillation. The residual error therefore reflects the approximation limit of $\Pi$ rather than a different target induced by privileged variation.
\end{corollary}

\begin{corollary}[Privilege use and distillable shift]
\label{cor:privilege-shift}
Let $A\sim q_U(\cdot\mid H)$. Then
\begin{equation}
    \mathbb E[D_{\mathrm{KL}}(q_U\Vert\overline q)]
    =I(A;U\mid H),
    \label{eq:irreducible-privilege}
\end{equation}
and, relative to the frozen student $p_k$,
\begin{equation}
    \mathbb E[D_{\mathrm{KL}}(q_U\Vert p_k)]
    =I(A;U\mid H)
    +\mathbb E[D_{\mathrm{KL}}(\overline q\Vert p_k)].
    \label{eq:privilege-shift-decomposition}
\end{equation}
\end{corollary}

Equation~(\ref{eq:privilege-shift-decomposition}) separates teacher change into privileged variation and the causal shift available to the student. Privileged variation can consume the entire teacher--student divergence without moving the student target.

\begin{proposition}[Perfect privileged teachers need not transfer]
\label{prop:perfect-teacher-no-transfer}
For every horizon $T\geq2$, there is a binary verification problem, a prefix-only student $p_k$, and a teacher that uses the student's complete trajectory such that the teacher succeeds with probability one, while $\overline q=p_k$ at every supervised prefix. Exact distillation leaves the student unchanged, and its success probability is $2^{1-T}$.
\end{proposition}

The gap can therefore approach one as the response grows. Teacher success alone gives no nontrivial lower bound on student improvement. Appendix~\ref{app:theory} gives the construction, in which the teacher uses a future student token to coordinate an otherwise successful response while every projected next-token target remains uniform.

We next identify a positive transfer condition. Let $Q_k(H,a)\in[0,1]$ be the probability of verifier success after taking action $a$ at $H$ and continuing with the frozen student. For any prefix-only distribution $r$, define its frozen-student value shift
\begin{equation}
    g_k(r;H)
    =\sum_a\bigl(r(a\mid H)-p_k(a\mid H)\bigr)Q_k(H,a).
    \label{eq:frozen-student-value-shift}
\end{equation}

\begin{proposition}[Value-aligned causal transfer]
\label{prop:value-aligned-transfer}
For almost every prefix $H$,
\begin{equation}
    g_k(\overline q;H)
    =\mathbb E[g_k(q_U;H)\mid H].
    \label{eq:value-survives-projection}
\end{equation}
Thus trajectory-specific changes may cancel in distribution, but any positive mean shift under the frozen student's value function survives causal projection.
\end{proposition}

The two outcome branches make this condition method-specific. Let $\eta(H)=P(R=0\mid H)$ and let $\overline q_+$ and $\overline q_-$ denote the conditional mean teachers on correct and failed trajectories. Suppose
\begin{equation}
\begin{split}
    \delta_+(H)
    &={\mathbb E}[D_{\mathrm{KL}}(p_k\Vert q_U)\mid H,R=1],\\
    \gamma_-(H)&=g_k(\overline q_-;H),
    \qquad
    \delta_S(H)=D_{\mathrm{KL}}(\overline q\Vert\pi).
\end{split}
\label{eq:transfer-margin-terms}
\end{equation}

\begin{corollary}[Outcome-conditioned transfer margin]
\label{cor:outcome-transfer-margin}
For any prefix-only student $\pi$,
\begin{equation}
    g_k(\pi;H)
    \geq
    \eta\gamma_-
    -(1-\eta)\sqrt{\delta_+/2}
    -\sqrt{\delta_S/2}.
    \label{eq:outcome-transfer-margin}
\end{equation}
The projected update has positive frozen-student value whenever the failed-branch margin exceeds the preservation and distillation errors on the right-hand side.
\end{corollary}

The correct branch makes $\delta_+=0$ at its ideal unclipped distributional optimum. A mean correct-branch objective at most $\delta$ implies mean total variation at most $\sqrt{\delta/2}$. The failed branch must supply $\gamma_->0$, not merely high privileged-teacher success. Its student-proximity term targets the compatibility needed for that distinction. Appendix~\ref{app:theory} proves a trajectory-level compatibility bound and connects the implemented success-path regularizer to this condition.

Finally, the value margin has a rollout-level implication. For the conservative interpolation $\pi_\epsilon=(1-\epsilon)p_k+\epsilon\pi$ in a finite-horizon problem,
\begin{equation}
    \left.\frac{d}{d\epsilon}J(\pi_\epsilon)\right|_{\epsilon=0}
    =\sum_t\mathbb E_{H_t\sim p_k}[g_k(\pi;H_t)].
    \label{eq:local-policy-improvement}
\end{equation}
If the right-hand side is positive, sufficiently small $\epsilon>0$ improves verifier success. The theorem identifies the transferable quantity and the positive margin required for local improvement. The implemented update targets this direction through pointwise clipping and a shared-parameter policy. The trajectory perturbations in Section~\ref{sec:privilege-diagnostics} test privilege use, while the factorial and objective studies test whether teacher adaptation produces a better causal student rather than only a more successful privileged teacher.
 \section{Experiments}
\label{sec:experiments}

The experiments test causal student improvement, factorial attribution, matching-trajectory use, outcome-branch contributions, objective robustness, sampling efficiency, and scale.

\subsection{Setup}
\label{sec:experimental-setup}

We train Qwen3-1.7B in thinking mode for 100 student updates with seeds 17, 29, and 43. Evaluation uses AIME 2024, AIME 2025, and HMMT 2025 at temperature 1.0 with 12 samples per problem. We report task Avg@12 and their equally weighted macro average. All methods share the student-update count and training protocol; confidence intervals use a paired bootstrap stratified by problem and seed. Appendix~\ref{app:experiments} gives the complete protocol and checks.

\subsection{PAST Improves the Causal Student}
\label{sec:main-results}

PAST produces the strongest final student in the comparison set. Its Avg@12 macro average is $48.426$, a $5.617$-point improvement over Vanilla OPSD with a 95\% confidence interval of $[2.592, 8.333]$. PAST also leads every task and exceeds H2SD and SD-Zero in macro average (Table~\ref{tab:main-results}).

\begin{table}[!htbp]
    \centering
    \caption{Final Avg@12 after 100 updates. PAST exceeds Vanilla OPSD by $5.617$ points (95\% CI $[2.592, 8.333]$).}
    \label{tab:main-results}
    \small
    \begin{tabular}{@{}lrrrr@{}}
        \toprule
        Method & AIME24 & AIME25 & HMMT25 & Macro \\
        \midrule
        Base & 53.056 & 34.907 & 27.870 & 38.611 \\
        SFT & 55.000 & 37.222 & 27.407 & 39.877 \\
        GRPO & 55.463 & 38.056 & 27.963 & 40.494 \\
        Vanilla OPSD & 57.130 & 39.722 & 31.574 & 42.809 \\
        SD-Zero & 59.444 & 42.315 & 30.185 & 43.981 \\
        H2SD & 61.667 & 42.500 & 32.130 & 45.432 \\
        \textbf{PAST} & \textbf{63.519} & \textbf{45.833} & \textbf{35.926} & \textbf{48.426} \\
        \bottomrule
    \end{tabular}
\end{table}

PAST's advantage develops over training. Its macro average rises at every checkpoint and progressively separates from Vanilla OPSD (Figure~\ref{fig:main-learning-dynamics}). The HMMT25 gain appears early; AIME24 and AIME25 vary more across checkpoints, but PAST finishes with the highest mean on all three tasks.

\subsection{Factorial Attribution of the PAST Gain}
\label{sec:factorial-attribution}

The $2\times2$ design separates trajectory access from teacher adaptation. The cell without either factor is Vanilla OPSD. Trajectory-only changes the teacher input; teacher-only adapts without the completed response; PAST combines the matching trajectory with adaptation.

Neither factor alone reproduces the gain. Trajectory access changes the teacher context but adds only $0.555$ macro points, while adaptation without the completed response adds $0.092$ points. Each condition lowers at least one task. PAST instead reaches $48.426$ and improves all three. This pattern supports complementary roles. The matching trajectory supplies student-specific hindsight, and adaptation converts it into a more useful student-facing target.

\begin{table}[!htbp]
    \centering
    \caption{PAST combines trajectory access and teacher adaptation to produce the strongest student.}
    \label{tab:factorial}
    \small
    \begin{tabular}{@{}ccrrrr@{}}
        \toprule
        Trajectory & Trained teacher & AIME24 & AIME25 & HMMT25 & Macro \\
        \midrule
        No  & No  & 57.130 & 39.722 & 31.574 & 42.809 \\
        No  & Yes & 59.074 & 38.056 & 31.574 & 42.901 \\
        Yes & No  & 59.537 & 41.111 & 29.444 & 43.364 \\
        Yes & Yes & \textbf{63.519} & \textbf{45.833} & \textbf{35.926} & \textbf{48.426} \\
        \bottomrule
    \end{tabular}
\end{table}

\subsection{The Teacher Uses Matching Trajectory Privilege}
\label{sec:privilege-diagnostics}

The teacher depends on the trajectory matching the current response. Teacher success is its continuation pass rate, while distribution JSD measures next-token change. Removing the trajectory tests student-specific context; cross-problem shuffling retains the field but breaks its correspondence with the problem and prefix, separating matched hindsight from an additional input alone.

Both perturbations weaken the teacher, with a larger effect from shuffling. Relative to the full condition, removal lowers success by $0.064$ and raises JSD by $0.067$; shuffling lowers success by $0.089$ and raises JSD by $0.110$ (Figure~\ref{fig:teacher-mechanism}a). In the E11 trace, correct-branch KL falls from $0.186$ to $0.077$ and failed-branch success rises from $0.410$ to $0.484$ (Figures~\ref{fig:teacher-mechanism}b and~\ref{fig:teacher-mechanism}c). These diagnostics establish matching-trajectory use and branch-level adaptation; Table~\ref{tab:factorial} separately establishes the final-student contribution.

\subsection{Joint Outcome-Conditioned Adaptation Performs Best}
\label{sec:branch-contribution}

Joint correct-and-failed training leads every task, reaching $48.426$ macro versus $45.864$ for failed-only and $44.908$ for correct-only training (Table~\ref{tab:branch-ablation}). Its $2.562$-point advantage over the stronger single branch supports joint adaptation.

\subsection{Objective Robustness, Sampling Efficiency, and Scale}
\label{sec:objective-robustness}

PAST remains effective across alternative distribution objectives and reduces teacher sampling cost, while the 4B extension shows task-dependent gains.

\FloatBarrier

Figure~\ref{fig:objective-tradeoffs} shows that generalized JSD reaches $47.623$, close to $47.562$ for full-vocabulary KL at slightly lower compatibility KL, while sampled-token KL trades lower quality for higher throughput. Adaptive sampling uses $17.6\%$ fewer teacher samples, reducing the count from 1,032 to 850, while reaching a slightly higher macro average of $47.284$ versus $46.728$ for fixed-size sampling (Table~\ref{tab:adaptive-sampling}). The controller reduces routine sampling once teacher success stabilizes while retaining extra draws for failed groups without a successful continuation (Figure~\ref{fig:adaptive-sampling}). On Qwen3-4B, PAST raises the macro average from $47.685$ to $48.519$, with gains on AIME25 and HMMT25 and a lower AIME24 score (Table~\ref{tab:scale-extension}). Appendices~\ref{app:additional-results} and~\ref{app:cost} give the complete ablations, learning curves, optimization diagnostics, 4B results, and training-cost accounting.
 \FloatBarrier
\section{Conclusion}
\label{sec:conclusion}

PAST uses completed on-policy responses as teacher-side privilege while retaining a prefix-only student interface. It preserves verified-correct behavior and adapts failed trajectories toward verifier success before distilling on the student's original prefixes. Our analysis identifies the causal projection of this future-conditioned teacher and conditions for improving verified success. Across three mathematical reasoning benchmarks, PAST improves Avg@12 over Vanilla OPSD by 5.6 points. Factorial and trajectory-perturbation studies attribute this gain to teacher adaptation with matching trajectories, showing how completed student experience can improve a causally deployed policy.

\label{page:main-content-end}
\clearpage

\section*{Limitations}

The main evidence uses a 1.7B student, while the one-seed 4B extension gives mixed task-level results. Broader scale behavior therefore remains open. The verifier checks final answers and does not validate each intermediate reasoning step. Reliable process feedback could extend the supervision signal within the same interface. The failed branch also uses success masking, adaptive group sizes, one retry, and all-failure skipping. Our theory covers the population objectives and causal projection rather than unbiasedness of this finite-sample estimator. The appendix reports skip and group-composition diagnostics that make this estimator boundary measurable.
 
\bibliography{refs}
\bibliographystyle{iclr2026_conference}
\clearpage

\appendix
\raggedbottom
\section{Related Work}
\label{sec:related-work}

\paragraph{On-policy distillation.}
On-policy knowledge distillation trains on student-generated states and reduces the mismatch between teacher-generated training sequences and student-generated inference states \citep{agarwal2024gkd,gu2024minillm}. DistiLLM studies stable divergence choices and efficient updates for this setting \citep{ko2024distillm}. On-policy self-distillation (OPSD) specializes the interface to reasoning, where a privileged teacher supplies full-vocabulary targets on an unprivileged student's rollout prefixes \citep{zhao2026opsd}. PAST retains these student-facing prefixes but changes how the teacher is formed. The completed student response becomes additional teacher privilege and trains the teacher before distillation.

\paragraph{Hindsight supervision for reasoning.}
Completed responses and outcome feedback support several forms of self-improvement. SD-Zero conditions a reviser on a completed response and its outcome before distilling the revision \citep{he2026sdzero}. H2SD applies different hindsight mechanisms to successful and failed responses \citep{cai2026h2sd}, while RSTG supplies adaptive teacher guidance to failed zero-variance reinforcement-learning groups \citep{han2026distillwherefail}. Other methods change the privileged representation or select supervision through problem-solving structure, partial solutions, position reliability, or token teachability \citep{zhao2026psopsd,tan2026paint,liu2026pwopsd,wang2026teachability}. PAST uses the completed response for a different operation. It adapts one OPSD teacher with preservation on correct trajectories and verifier-seeking, student-regularized optimization on failed trajectories, then distills that teacher on the original student prefixes.

\paragraph{Privileged and trainable teachers.}
Learning using privileged information studies variables available during training but absent at inference \citep{vapnik2009lupi}, and generalized distillation connects this setting to teacher-student learning \citep{lopezpaz2016generalizeddistillation}. Trainable teachers provide a complementary direction. For example, OKD adapts online teacher modules during autoregressive distillation \citep{rao2026okd}. PAST combines teacher adaptation with a specific information constraint. Its teacher observes the student's complete future trajectory during training, while the deployed student remains prefix-only. Section~\ref{sec:theory} characterizes the part of this future-dependent teaching distribution that survives causal projection.

 \section{Method and Protocol Details}
\label{app:method}

\subsection{Comparison of Student-Facing Interfaces}

Table~\ref{tab:distillation-interface-comparison} separates the source of the teacher distribution from the states used for the student update. All listed distillation methods train on prefixes sampled from the student. Their main differences lie in teacher construction, privileged context, and the role of outcome feedback.

\begin{table}[t]
    \centering
    \caption{Comparison of student-facing training interfaces. PAST uses the completed response to adapt the teacher while retaining the original student prefixes for distillation.}
    \label{tab:distillation-interface-comparison}
    \scriptsize
    \setlength{\tabcolsep}{3pt}
    \begin{tabular}{@{}p{0.09\linewidth}p{0.16\linewidth}p{0.24\linewidth}p{0.18\linewidth}p{0.19\linewidth}@{}}
        \toprule
        Method & Teacher construction & Teacher context beyond the prefix & Outcome role & Student update states \\
        \midrule
        OPD \citep{agarwal2024gkd,gu2024minillm}
        & Fixed external teacher
        & No additional privilege
        & Not required
        & Student prefixes \\
        OPSD \citep{zhao2026opsd}
        & Stop-gradient privileged self-teacher
        & Task-side privilege $Z$
        & Not required
        & Student prefixes \\
        SD-Zero \citep{he2026sdzero}
        & Self-revision-trained teacher, frozen during distillation
        & Complete response and outcome-conditioned instruction
        & Enters the teacher context
        & Student prefixes \\
        H2SD \citep{cai2026h2sd}
        & Fixed self-teacher with offline hints
        & Verified response or external hint
        & Selects the success or failure objective
        & Student prefixes \\
        PAST
        & Cycle-adapted privileged self-teacher
        & Task privilege $Z$ and complete response $Y$
        & Selects teacher adaptation but is excluded from $U$
        & Original student prefixes \\
        \bottomrule
    \end{tabular}
\end{table}

\subsection{Teacher Input Serialization}

All teacher phases use one user message and the model's native chat template. The serialized fields appear in the following order:
\begin{verbatim}
Problem: {problem}

Please reason step by step, and put your final answer within \boxed{}.

=== Student Attempt Begin ===
{student_output}
=== Student Attempt End ===

=== Reference Solution Begin ===
{reference_solution}
=== Reference Solution End ===

The student attempt above may be correct or incorrect. Use it as hindsight
context for the student's reasoning state. Maintain the student's established
reasoning style and presentation whenever they are compatible with a correct
solution. Use the reference solution to ensure correctness, but do not copy or
paraphrase the reference solution. Now solve the original problem through your
own reasoning, and put the final answer within \boxed{}.
\end{verbatim}
The student attempt is always produced by the current unprivileged student. Correct and failed trajectories use the same message; correctness is kept outside the prompt and controls only routing and verification. In the correct branch and student distillation phase, the assistant segment is teacher-forced on the original student response. In the failed branch, the teacher samples the assistant segment autoregressively.

\subsection{Adaptive Failed-Branch Sampling}

Let $G_{\max}$ be the fixed group size of the GRPO baseline and let $G_{\mathrm{base}}\in[1,G_{\max}]$ be the current base draw. Warm-up starts at $G_{\mathrm{base}}=G_{\max}$. Once the recent teacher success rate stabilizes above the configured threshold, the controller decreases $G_{\mathrm{base}}$ one step at a time, down to one. For each failed student trajectory, PAST applies the following state machine.
\begin{enumerate}
    \item Draw $G_{\mathrm{base}}$ teacher continuations.
    \item If at least one succeeds, classify the realized group as mixed or all-success and compute Eq.~(\ref{eq:failed-teacher-loss}).
    \item If all fail, add one continuation at a time until the first success or until the group reaches $G_{\max}$.
    \item If a size-$G_{\max}$ group remains all-failure, redraw one complete size-$G_{\max}$ group. Skip the trajectory only when this retry also contains no success.
\end{enumerate}
A mixed group uses relative advantages and successful-path KL. An all-success group, including the single-success case, has zero relative advantage and uses only successful-path KL. An all-failure group contributes neither GRPO nor student-proximity KL.

\subsection{Loss Reduction and Clipping Diagnostics}

For $N_+$ verified-correct trajectories and an active failed-group set $\mathcal G_-^{\mathrm{act}}$, the branch means are
\begin{equation}
    \overline{\mathcal L}_+
    =\frac{1}{N_+}\sum_{i:R_i=1}\mathcal L_{+,i},
    \qquad
    \overline{\mathcal L}_-
    =\frac{1}{|\mathcal G_-^{\mathrm{act}}|}
    \sum_{g\in\mathcal G_-^{\mathrm{act}}}\mathcal L_{-,g}.
\end{equation}
Only existing means enter Eq.~(\ref{eq:teacher-active-branch-loss}); the optimizer step is skipped when neither branch is active. Correct-teacher and student-distillation losses clip individual vocabulary contributions at the Vanilla OPSD threshold. Failed-path KL remains exact. Training logs retain the clipped loss, unclipped KL, clip fraction, removed positive mass, failed-path KL quantiles, and non-finite counts.

\subsection{Student and Teacher Snapshot Lifecycle}

PAST uses one frozen base model with separate student and teacher LoRA adapters. At the start of cycle $k$, the student adapter is the read-only policy $p_k$ and the teacher adapter is $q_k$. The student adapter generates the rollout; only the teacher adapter is trainable during teacher adaptation. PAST then freezes $q_{k+1}$ and makes the student adapter trainable for one logical update from $p_k$ to $p_{k+1}$. Gradient accumulation may split either logical update into microbatches but does not add optimizer steps or replay rollouts.

Checkpoints are written only at complete cycle boundaries and contain both adapters, both optimizer and scheduler states, the adaptive-sampling controller, data position, random-number-generator states, and configuration identifiers. An interrupted partial cycle is discarded and restarted from the preceding boundary. Full-vocabulary reference logits are recomputed under the appropriate frozen adapter and released after each microbatch rather than stored across phases.

 \section{Complete Theory and Proofs}
\label{app:theory}

\subsection{Conditions and Conditioning Measure}

All expectations in Section~\ref{sec:theory} are taken under the joint distribution induced by the task distribution, the frozen student's on-policy rollout, the prefix-selection rule, and the adapted teacher's next-token distribution. For almost every prefix $H$, the conditional teacher distributions $q_U(\cdot\mid H)$ are defined on the common vocabulary $\mathcal V$. We assume that the displayed KL divergences and conditional expectations are finite. Uniqueness holds on actions with positive mass under the corresponding target distribution. These conditions let conditional expectation and finite-vocabulary summation be interchanged below.

The value results additionally assume a finite-horizon generation problem with terminal verifier reward in $[0,1]$. For a frozen student $p_k$, $Q_k(H,a)$ is the expected terminal reward after taking $a$ at $H$ and following $p_k$ thereafter. Conditional branch means are needed only when the corresponding event has positive probability. Terms multiplied by a zero branch probability are defined as zero.

\subsection{Proof of the Causal Projection}

\begin{proof}[Proof of Proposition~\ref{prop:causal-projection}]
For any prefix-only $\pi$, insert $\overline q$ into the log ratio
\begin{equation}
    \log\frac{q_U(A\mid H)}{\pi(A\mid H)}
    =\log\frac{q_U(A\mid H)}{\overline q(A\mid H)}
    +\log\frac{\overline q(A\mid H)}{\pi(A\mid H)}.
\end{equation}
Taking expectations gives the first term in Eq.~(\ref{eq:projection-decomposition}). For the second term, conditioning on $H$ replaces $q_U(a\mid H)$ by $\mathbb E[q_U(a\mid H)\mid H]=\overline q(a\mid H)$, which yields $\mathbb E[D_{\mathrm{KL}}(\overline q\Vert\pi)]$. Nonnegativity establishes optimality, and equality requires $\pi=\overline q$ wherever $\overline q$ has positive support.
\end{proof}

\begin{proof}[Proof of Corollary~\ref{cor:restricted-causal-projection}]
Proposition~\ref{prop:causal-projection} writes the privileged-teacher objective as
\[
\mathbb E[D_{\mathrm{KL}}(q_U\Vert\pi)]
=
\mathbb E[D_{\mathrm{KL}}(q_U\Vert\overline q)]
+
\mathbb E[D_{\mathrm{KL}}(\overline q\Vert\pi)].
\]
The first term is independent of $\pi$. Subtracting the infimum over $\rho\in\Pi$ gives Eq.~(\ref{eq:restricted-projection-excess}). The excess risks are identical pointwise, so the objectives have the same minimizing sequences and the same minimizers whenever a minimum is attained. If $\overline q\in\Pi$, the remaining KL is minimized uniquely by $\overline q$ on its support.
\end{proof}

\begin{proof}[Proof of Corollary~\ref{cor:privilege-shift}]
Under the joint law $P(A,U\mid H)=P(U\mid H)q_U(A\mid H)$, the conditional marginal of $A$ is $\overline q(A\mid H)$. The definition of conditional mutual information gives Eq.~(\ref{eq:irreducible-privilege}). Applying Proposition~\ref{prop:causal-projection} with $\pi=p_k$ gives Eq.~(\ref{eq:privilege-shift-decomposition}).
\end{proof}

Whenever the expected teacher--student KL is positive, Eq.~(\ref{eq:privilege-shift-decomposition}) also gives the descriptive fraction
\begin{equation}
    \frac{\mathbb E[D_{\mathrm{KL}}(\overline q\Vert p_k)]}
    {\mathbb E[D_{\mathrm{KL}}(q_U\Vert p_k)]}
    =1-
    \frac{I(A;U\mid H)}
    {\mathbb E[D_{\mathrm{KL}}(q_U\Vert p_k)]}.
    \label{eq:surviving-shift-fraction}
\end{equation}
This ratio is an interpretation of the decomposition, not an objective optimized explicitly by PAST.

\subsection{A Perfect Privileged Teacher Need Not Transfer}

\begin{proof}[Proof of Proposition~\ref{prop:perfect-teacher-no-transfer}]
Fix $T\geq2$. A response is a binary string $C=(C_1,\ldots,C_T)$, and the verifier accepts exactly the two constant strings
\begin{equation}
    V(C)=\mathbf{1}\{C_1=\cdots=C_T\}.
\end{equation}
Let the frozen student generate independent fair bits. Its success probability is therefore $2/2^T=2^{1-T}$.

Draw the student's complete training trajectory $Y\sim p_k$ and include it in $U$. Let $B=Y_T$, which is available to the privileged teacher but is absent from every student prefix $H_t=Y_{<t}$. Define the teacher to output $B$ deterministically at every position,
\begin{equation}
    q_U(a\mid H_t)=\mathbf{1}\{a=B\}.
\end{equation}
When this teacher generates a response under the fixed privilege $U$, it produces $(B,\ldots,B)$ and succeeds with probability one.

For every supervised position $t\leq T$, the fair bit $B=Y_T$ is independent of $H_t=Y_{<t}$. Hence
\begin{equation}
    \overline q(a\mid H_t)
    =\mathbb E[\mathbf{1}\{a=B\}\mid H_t]
    =\tfrac12
    =p_k(a\mid H_t).
\end{equation}
The exact forward-KL projection is the original student at every supervised prefix, so distillation makes no update. Moreover, $A=B$ under the teacher and $B$ remains uniform given $H_t$, which gives $I(A;U\mid H_t)=\log 2$ while $D_{\mathrm{KL}}(\overline q\Vert p_k)=0$. The privileged teacher is perfect, yet its advantage over the causal student approaches one as $T$ grows.
\end{proof}

The construction uses the future trajectory as a coordination variable. Its success is a property of repeatedly conditioning on the same unavailable bit, not of a next-token change that a prefix-only student can retain.

\subsection{Value-Aligned Causal Transfer}

\begin{proof}[Proof of Proposition~\ref{prop:value-aligned-transfer}]
Condition on $H$. Since $Q_k(H,a)$ and $p_k(a\mid H)$ do not depend on $U$,
\begin{align}
\mathbb E[g_k(q_U;H)\mid H]
&=\sum_a\left(\mathbb E[q_U(a\mid H)\mid H]-p_k(a\mid H)\right)Q_k(H,a)\\
&=\sum_a\left(\overline q(a\mid H)-p_k(a\mid H)\right)Q_k(H,a)\\
&=g_k(\overline q;H).
\end{align}
Thus conditional averaging preserves the mean of every bounded prefix-action value shared across privilege conditions. Distributional cancellation erases value only when positive and negative value shifts cancel as well.
\end{proof}

Let $\eta=P(R=0\mid H)$. The conditional barycenter obeys
\begin{equation}
    \overline q
    =(1-\eta)\overline q_++\eta\overline q_-,
    \label{eq:outcome-barycenter}
\end{equation}
and linearity gives
\begin{equation}
    g_k(\overline q;H)
    =(1-\eta)g_k(\overline q_+;H)
    +\eta g_k(\overline q_-;H).
    \label{eq:outcome-value-decomposition}
\end{equation}
At the ideal correct-branch optimum $\overline q_+=p_k$, the full projected shift is $\overline q-p_k=\eta(\overline q_--p_k)$. In particular,
\begin{equation}
    \operatorname{TV}(\overline q,p_k)
    =\eta\operatorname{TV}(\overline q_-,p_k).
    \label{eq:failure-tv-scaling}
\end{equation}
Failed-branch signal is therefore attenuated by its conditional frequency even before student approximation error is introduced.

\begin{proof}[Proof of Corollary~\ref{cor:outcome-transfer-margin}]
Because $Q_k(H,\cdot)\in[0,1]$, the difference in its expectation under two action distributions is at most their total variation. Convexity of total variation, Pinsker's inequality, and Jensen's inequality give
\begin{align}
g_k(\overline q_+;H)
&\geq-\operatorname{TV}(\overline q_+,p_k)\\
&\geq-\mathbb E[\operatorname{TV}(q_U,p_k)\mid H,R=1]\\
&\geq-\sqrt{\delta_+(H)/2}.
\end{align}
Using Eq.~(\ref{eq:outcome-value-decomposition}) and $\gamma_-=g_k(\overline q_-;H)$ yields
\begin{equation}
    g_k(\overline q;H)
    \geq\eta\gamma_--(1-\eta)\sqrt{\delta_+/2}.
\end{equation}
The same bounded-value argument and Pinsker's inequality give
\begin{equation}
    |g_k(\pi;H)-g_k(\overline q;H)|
    \leq\operatorname{TV}(\pi,\overline q)
    \leq\sqrt{D_{\mathrm{KL}}(\overline q\Vert\pi)/2}.
\end{equation}
Substituting $\delta_S=D_{\mathrm{KL}}(\overline q\Vert\pi)$ proves Eq.~(\ref{eq:outcome-transfer-margin}).
\end{proof}

\subsection{Compatibility Between Privileged and Student Continuations}

The failed teacher is evaluated under its own privileged continuation, whereas $g_k$ evaluates the teacher's next-token distribution followed by the frozen student. A continuation-distance condition connects these quantities.

\begin{proposition}[Continuation compatibility]
\label{prop:continuation-compatibility}
Fix $(H,U)$. Let $\mathbb Q_U$ be the distribution of a complete continuation generated by $q_U$. Let $\mathbb M_U$ draw its first action from $q_U(\cdot\mid H)$ and all later actions from $p_k$. Write $s_U=\mathbb E_{\mathbb Q_U}[V]$ and $v_k=\mathbb E_{p_k}[V\mid H]$. Then
\begin{equation}
    g_k(q_U;H)
    \geq s_U-v_k
    -\sqrt{D_{\mathrm{KL}}(\mathbb Q_U\Vert\mathbb M_U)/2}.
    \label{eq:compatibility-value-bound}
\end{equation}
Moreover, the trajectory KL has the chain-rule form
\begin{equation}
    D_{\mathrm{KL}}(\mathbb Q_U\Vert\mathbb M_U)
    =\mathbb E_{C\sim\mathbb Q_U}
    \left[\sum_{s>t}
    D_{\mathrm{KL}}\!\left(
        q_U(\cdot\mid G_s)
        \Vert p_k(\cdot\mid G_s)
    \right)\right],
    \label{eq:continuation-chain-rule-kl}
\end{equation}
where $t$ is the position following $H$ and $G_s$ is the generated prefix at position $s$.
\end{proposition}

\begin{proof}
The hybrid success probability is
\begin{equation}
    \mathbb E_{\mathbb M_U}[V]
    =\sum_a q_U(a\mid H)Q_k(H,a)
    =v_k+g_k(q_U;H).
\end{equation}
Since $V\in[0,1]$, its expectation differs under $\mathbb Q_U$ and $\mathbb M_U$ by at most their total variation. Pinsker's inequality gives
\begin{equation}
    s_U-\mathbb E_{\mathbb M_U}[V]
    \leq\operatorname{TV}(\mathbb Q_U,\mathbb M_U)
    \leq\sqrt{D_{\mathrm{KL}}(\mathbb Q_U\Vert\mathbb M_U)/2},
\end{equation}
which proves Eq.~(\ref{eq:compatibility-value-bound}). The two trajectory laws use the same first-action distribution. Applying the KL chain rule to all later positions gives Eq.~(\ref{eq:continuation-chain-rule-kl}).
\end{proof}

Proposition~\ref{prop:continuation-compatibility} formalizes the role of student proximity. High privileged success gives a positive compatible value margin only when the teacher's advantage exceeds its continuation mismatch. The implemented quantity in Eq.~(\ref{eq:failed-path-kl}) evaluates the same tokenwise divergence on sampled verifier-successful paths. Because that loss is success-conditioned rather than the unconditional trajectory KL in Eq.~(\ref{eq:continuation-chain-rule-kl}), it encourages compatibility but does not certify this bound globally.

\subsection{Local Rollout Improvement}

\begin{proposition}[Local policy improvement]
\label{prop:local-policy-improvement}
Let $J(r)$ denote expected terminal verifier reward under a finite-horizon prefix-only policy $r$. For any prefix-only $\pi$, define $\pi_\epsilon=(1-\epsilon)p_k+\epsilon\pi$. Then Eq.~(\ref{eq:local-policy-improvement}) holds. If its right-hand side is strictly positive, there exists $\epsilon_0>0$ such that $J(\pi_\epsilon)>J(p_k)$ for every $\epsilon\in(0,\epsilon_0)$.
\end{proposition}

\begin{proof}
Because the horizon and vocabulary are finite, $J(\pi_\epsilon)$ is a finite sum of products of action probabilities and is differentiable at $\epsilon=0$. Differentiating one policy factor at a time leaves $p_k$ on every preceding and subsequent factor. The contribution at position $t$ is
\begin{equation}
    \mathbb E_{H_t\sim p_k}
    \left[\sum_a
    \bigl(\pi(a\mid H_t)-p_k(a\mid H_t)\bigr)
    Q_k(H_t,a)\right]
    =\mathbb E_{H_t\sim p_k}[g_k(\pi;H_t)].
\end{equation}
Summing over positions proves Eq.~(\ref{eq:local-policy-improvement}). A differentiable function with a strictly positive derivative at zero is larger than its value at zero throughout some sufficiently small right neighborhood, which proves the second claim.
\end{proof}

This result concerns a conservative policy-space interpolation. It does not identify a finite parameter-space gradient step with $\pi_\epsilon$, nor does it establish improvement when the occupancy-weighted margin is nonpositive.

\subsection{Forward and Reverse Projection}

Forward and reverse KL aggregate privileged targets differently. Proposition~\ref{prop:causal-projection} gives the arithmetic barycenter
\begin{equation}
    \argmin_\pi \mathbb E[D_{\mathrm{KL}}(q_U\Vert\pi)]
    =\mathbb E[q_U\mid H].
\end{equation}
For reverse KL, direct Lagrangian optimization gives
\begin{equation}
    \argmin_\pi \mathbb E[D_{\mathrm{KL}}(\pi\Vert q_U)]
    \propto \exp\!\left(\mathbb E[\log q_U\mid H]\right).
\end{equation}
PAST uses the divergence direction that preserves the conditional mean soft target. Reverse KL instead suppresses actions that are not supported consistently across hindsight conditions.

\subsection{Correct-Branch Stability and the Clipping Boundary}

\begin{proposition}[Correct trajectories are ideal fixed points]
\label{prop:correct-fixed-point}
On verified-correct trajectories, consider the unclipped population objective
\begin{equation}
    \mathcal L_+(q)
    =\mathbb E[D_{\mathrm{KL}}(p_k\Vert q)\mid R=1].
\end{equation}
Its distributional optimum is $q=p_k$ on the supported prefixes. If $\mathcal L_+(q)\leq\delta$, then
\begin{equation}
    \mathbb E[\operatorname{TV}(p_k,q)\mid R=1]
    \leq\sqrt{\delta/2}.
    \label{eq:correct-tv-bound}
\end{equation}
\end{proposition}

\begin{proof}[Proof of Proposition~\ref{prop:correct-fixed-point}]
Nonnegativity of KL gives the unique supported distributional optimum $q=p_k$. For the approximate statement, Pinsker's inequality gives $\operatorname{TV}(p_k,q)\leq\sqrt{D_{\mathrm{KL}}(p_k\Vert q)/2}$ at each condition. Jensen's inequality and the assumed mean KL bound yield Eq.~(\ref{eq:correct-tv-bound}).
\end{proof}

For a vocabulary distribution $p$ and threshold $\tau>0$, the pointwise-clipped objective in Eq.~(\ref{eq:pointwise-clipped-divergence}) equals the true KL whenever
\begin{equation}
    q(v)\geq p(v)\exp[-\tau/p(v)]
    \quad\text{for every }v\text{ with }p(v)>0.
    \label{eq:clipping-safe-region}
\end{equation}
The equality point $q=p$ lies inside this region, so the clipped and unclipped objectives have the same local gradient and Hessian around the correct fixed point. The clipped quantity is not a global divergence. For example, with $p=(0.9,0.1)$, $q=(0.001,0.999)$, and $\tau=0.05$, the true KL is $5.892$, whereas the clipped sum is $-0.180<0$. No global KL claim in this paper uses the clipped objective.

\subsection{Adaptive-Sampling Identities}

Suppose independent teacher continuations succeed with probability $s\in(0,1)$ and the maximum group size is $G$. A fixed group is all-failure with probability $(1-s)^G$, all-success with probability $s^G$, and mixed otherwise. If PAST begins with $b$ samples and adds one sample after each all-failure prefix, the expected number of first-stage samples is
\begin{equation}
    \mathbb E[N_1]
    =b+\sum_{j=b}^{G-1}(1-s)^j.
\end{equation}
When a complete all-failure group triggers one independent size-$G$ retry, the total expected sample count and final skip probability are
\begin{equation}
    \mathbb E[N]
    =b+\sum_{j=b}^{G-1}(1-s)^j+G(1-s)^G,
    \qquad
    P(\mathrm{skip})=(1-s)^{2G}.
\end{equation}
These identities explain the sampling reduction but do not make the success-masked, retry-conditioned GRPO estimator unbiased for an unconditional population objective.

 \section{Full Experimental Protocol}
\label{app:experiments}

\subsection{Training and Evaluation}

The main experiments use Qwen3-1.7B in thinking mode. Each method receives 100 logical student updates and is evaluated at checkpoints 0, 25, 50, 75, and 100. Training seeds are 17, 29, and 43. Evaluation covers AIME 2024, AIME 2025, and HMMT 2025 with temperature 1.0 and 12 generations per problem. Avg@12 is computed within each task, and the macro average weights the three tasks equally. The primary interval resamples paired observations within problem and training-seed strata.

The verifier parses the boxed final answer and compares it with the authorized task answer. Truncated, malformed, or unparseable responses are verifier failures. The resulting signal evaluates final-answer correctness rather than intermediate reasoning. All main comparisons share the student-update count and evaluation protocol. Table~\ref{tab:cost-accounting} reports teacher sampling costs separately.

\subsection{Protocol and Semantic Checks}

Table~\ref{tab:protocol-checks} summarizes repeated checks performed before the main run, including the observed end-to-end parity difference alongside semantic, numerical, and recovery checks.

\begin{table}[!tbp]
    \centering
    \caption{Protocol, numerical, and recovery checks, each repeated eight times.}
    \label{tab:protocol-checks}
    \small
    \begin{tabular}{@{}lrrr@{}}
        \toprule
        Check & Observed mean & Passed & Repeats \\
        \midrule
        Model and thinking mode & 1.000000 & 8 & 8 \\
        Training-data protocol & 1.000000 & 8 & 8 \\
        Evaluation protocol & 1.000000 & 8 & 8 \\
        Batch-loss absolute error & 0.000000 & 8 & 8 \\
        Batch-gradient cosine & 0.999989 & 8 & 8 \\
        End-to-end paired difference & -0.102349 & 8 & 8 \\
        CPU semantic tests & 1.000000 & 8 & 8 \\
        Resume and RNG recovery & 1.000000 & 8 & 8 \\
        \bottomrule
    \end{tabular}
\end{table}

\subsection{Full-Vocabulary Implementation}

Full-vocabulary KL is computed in PyTorch under the corresponding frozen adapter. Table~\ref{tab:full-vocab-implementation} compares the training implementation with an FP32 reference. The main BF16 path has gradient cosine $0.999457$ and no non-finite events. Sequence chunking and the fused kernel are implementation alternatives; they do not change the stated objective.

\begin{table}[!tbp]
    \centering
    \caption{Numerical accuracy and system measurements for full-vocabulary KL implementations.}
    \label{tab:full-vocab-implementation}
    \small
    \begin{tabular}{@{}lrrrrr@{}}
        \toprule
        Implementation & Loss error & Grad. cosine & GB & Tokens/s & Non-finite \\
        \midrule
        FP32 reference & 0.000000 & 0.999986 & 72.305 & 1854 & 0 \\
        BF16 full vocabulary & 0.002978 & 0.999457 & 61.718 & 3306 & 0 \\
        Sequence chunking & 0.002525 & 0.999558 & 49.967 & 2881 & 0 \\
        Fused kernel & 0.003780 & 0.999196 & 55.156 & 3709 & 0 \\
        \bottomrule
    \end{tabular}
\end{table}

The system records token IDs, masks, response lengths, verifier outcomes, failure types, cycle identifiers, and sampled-action log probabilities. Full-vocabulary reference distributions are recomputed for each microbatch and are not persisted across phases. Checkpoint recovery restores both adapters, optimizers, schedulers, sampling-controller state, data position, and random-number-generator states at a complete cycle boundary.
 \section{Additional Results and Diagnostics}
\label{app:additional-results}

\subsection{Learning Curves}

\begin{figure}[!htbp]
    \centering
    \includegraphics[width=0.88\linewidth]{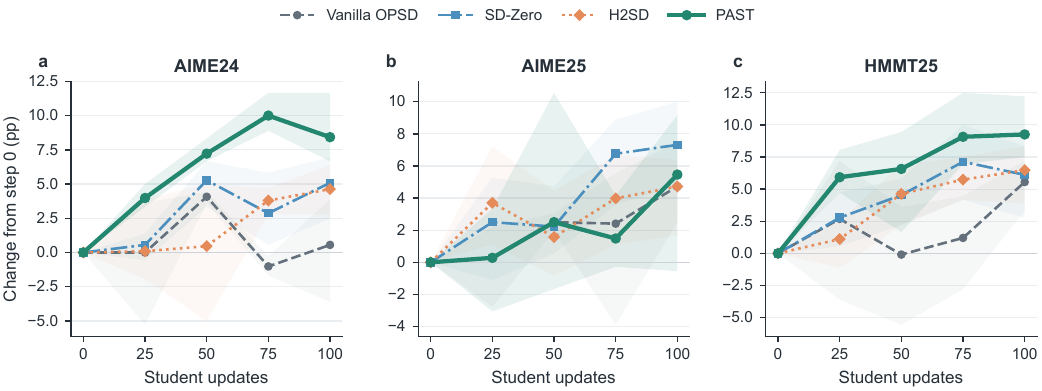}
    \caption{Avg@12 change from each method's own checkpoint 0. Lines show the mean over three training seeds and shaded regions show the seed range.}
    \label{fig:learning-curves}
\end{figure}

Figure~\ref{fig:learning-curves} reports changes from each method's own checkpoint-0 evaluation. This normalization isolates within-run learning and is not used for the absolute comparison in Table~\ref{tab:main-results}.

\subsection{Correct and Failed Branches}

Joint training has the strongest point estimate in every task and improves the macro average by $2.562$ points over the stronger single branch (Table~\ref{tab:branch-ablation}).

\begin{table}[!htbp]
    \centering
    \caption{Correct-only, failed-only, and joint teacher adaptation.}
    \label{tab:branch-ablation}
    \small
    \begin{tabular}{@{}lrrrr@{}}
        \toprule
        Training branch & AIME24 & AIME25 & HMMT25 & Macro \\
        \midrule
        Correct only & 60.093 & 43.611 & 31.019 & 44.908 \\
        Failed only & 62.037 & 43.333 & 32.222 & 45.864 \\
        Full PAST & \textbf{63.519} & \textbf{45.833} & \textbf{35.926} & \textbf{48.426} \\
        \bottomrule
    \end{tabular}
\end{table}

\begin{figure}[!htbp]
    \centering
    \includegraphics[width=0.58\linewidth]{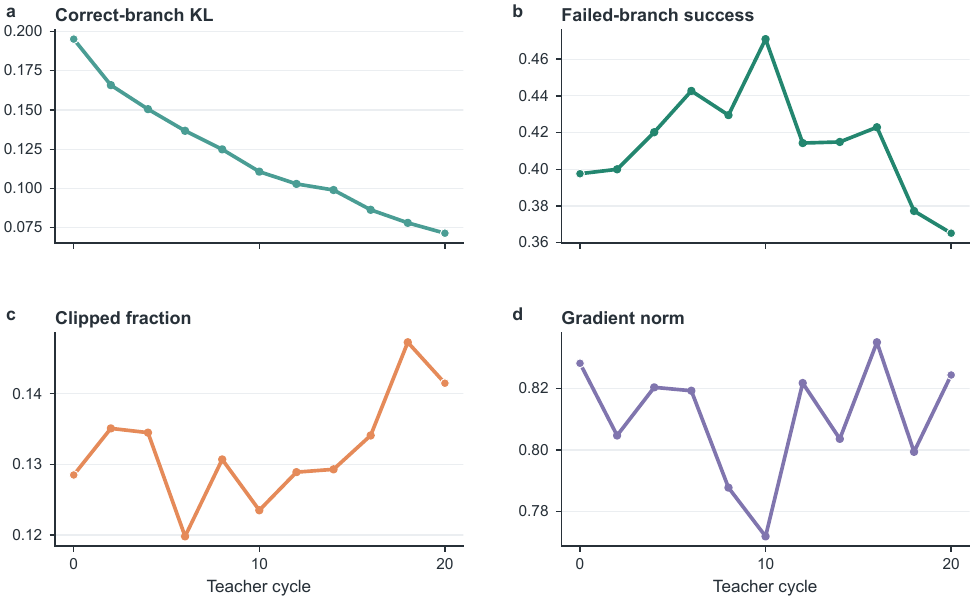}
    \caption{Teacher-training dynamics in E04. Correct-branch KL decreases, while failed-branch success and optimization diagnostics vary across cycles.}
    \label{fig:teacher-dynamics}
\end{figure}

Figure~\ref{fig:teacher-dynamics} shows that the correct-branch KL decreases from $0.1951$ to $0.0715$. Failed-branch success peaks at $0.4709$ and ends at $0.3652$, showing that success can fluctuate even while the preservation loss decreases; the clipping fraction and gradient norm remain finite throughout this trace.

\FloatBarrier

\subsection{Distribution Objectives}

The full-vocabulary reference and generalized JSD give similar one-seed quality, whereas sampled-token KL is cheaper but lower in this run. The failed-path regularization coefficient controls a visible success--compatibility trade-off: removing it raises teacher success but also produces the largest compatibility KL. These comparisons support the full-vocabulary objective as a strong quality--compatibility operating point.

\begin{table}[!htbp]
    \centering
    \caption{One-seed objective and regularization ablations. Compatibility is measured by KL to the frozen student.}
    \label{tab:objective-ablation}
    \scriptsize
    \begin{tabular}{@{}lrrrrr@{}}
        \toprule
        Objective & Macro & Teacher success & Compat. KL & GB & Tokens/s \\
        \midrule
        Full-vocabulary reference & 47.562 & 0.7323 & 0.0961 & 61.741 & 3311 \\
        Sampled-token KL & 45.864 & 0.7294 & 0.1220 & 43.804 & 4636 \\
        Generalized JSD & \textbf{47.623} & 0.7169 & 0.0877 & 61.949 & 3175 \\
        KL + CE & 44.815 & 0.7447 & 0.1326 & 60.909 & 3373 \\
        Top-$k$ approximation & 43.519 & 0.7223 & 0.1140 & 48.069 & 4199 \\
        $\beta_{\mathrm{KL}}=0$ & 43.395 & \textbf{0.7855} & 0.1722 & 61.390 & 3331 \\
        $\beta_{\mathrm{KL}}=0.01$ & 44.259 & 0.7480 & 0.1346 & 61.781 & 3293 \\
        $\beta_{\mathrm{KL}}=0.1$ & 43.735 & 0.6984 & 0.0744 & 62.005 & 3265 \\
        Failed-path clipping & 43.981 & 0.6823 & 0.0828 & 61.637 & 3281 \\
        Correct clipping off & 45.000 & 0.7517 & 0.1526 & 61.857 & 3359 \\
        \bottomrule
    \end{tabular}
\end{table}

\subsection{Adaptive Sampling}

Adaptive sampling reduces teacher generation while reaching a slightly higher observed macro average. It uses 850 teacher samples and 743,366 generated tokens, compared with 1,032 samples and 862,342 tokens for fixed-size sampling.

\begin{table}[!htbp]
    \centering
    \caption{Adaptive versus fixed-size teacher sampling.}
    \label{tab:adaptive-sampling}
    \small
    \begin{tabular}{@{}lrrrr@{}}
        \toprule
        Sampler & Macro & Samples & Tokens & GPU-hours \\
        \midrule
        Adaptive & \textbf{47.284} & \textbf{850} & \textbf{743,366} & \textbf{0.0392} \\
        Fixed-8 & 46.728 & 1,032 & 862,342 & 0.0457 \\
        \bottomrule
    \end{tabular}
\end{table}

\FloatBarrier

\subsection{Privilege and Optimization Diagnostics}

\begin{figure}[!htbp]
    \centering
    \includegraphics[width=0.62\linewidth]{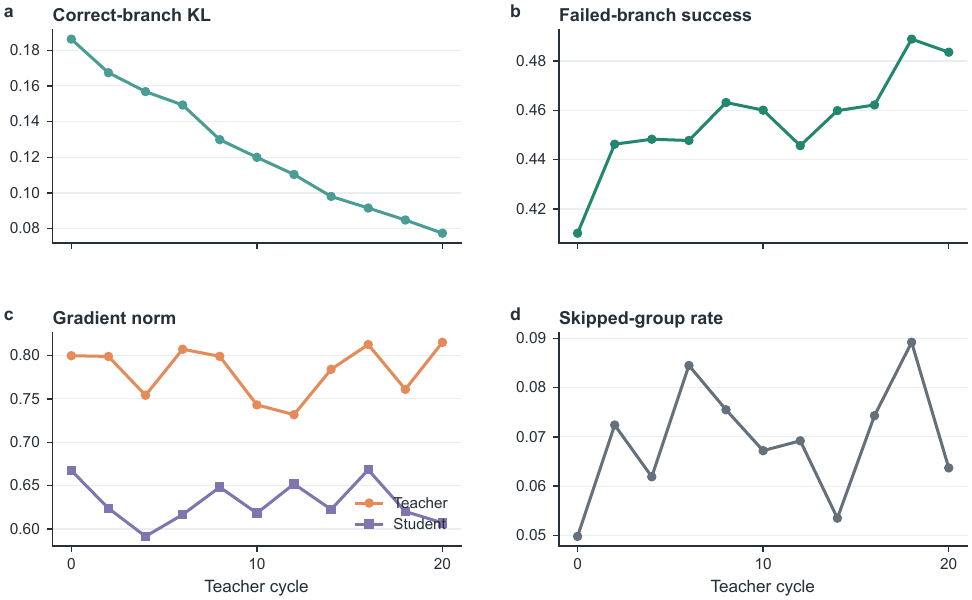}
    \caption{Cycle-level correct-branch KL, failed-branch success, gradient norms, and skipped-group rate.}
    \label{fig:privilege-dynamics}
\end{figure}

Figure~\ref{fig:privilege-dynamics} complements the condition comparison in Figure~\ref{fig:teacher-mechanism} with cycle-level optimization traces. In this E11 trace, correct-branch KL decreases from $0.1862$ to $0.0775$ and failed-branch success rises from $0.4102$ to $0.4836$. The skipped-group rate quantifies how often the finite-sample estimator lacks an active failed-branch update.

\subsection{One-Seed 4B Extension}

PAST has the highest macro point estimate in the one-seed Qwen3-4B extension. It improves AIME25 and HMMT25 over Vanilla OPSD, while Vanilla OPSD remains stronger on AIME24 (Table~\ref{tab:scale-extension}). This task-level variation motivates a broader multi-seed scale study.

\begin{table}[!htbp]
    \centering
    \caption{One-seed Qwen3-4B extension.}
    \label{tab:scale-extension}
    \small
    \begin{tabular}{@{}lrrrr@{}}
        \toprule
        Method & AIME24 & AIME25 & HMMT25 & Macro \\
        \midrule
        Vanilla OPSD & \textbf{66.111} & 45.000 & 31.944 & 47.685 \\
        Trajectory input only & 64.444 & 42.222 & \textbf{34.444} & 47.037 \\
        PAST & 62.778 & \textbf{48.611} & 34.167 & \textbf{48.519} \\
        \bottomrule
    \end{tabular}
\end{table}

\FloatBarrier
 \section{Cost and Reproducibility}
\label{app:cost}

\subsection{Training and Deployment Cost}

Table~\ref{tab:cost-accounting} reports training cost through generated teacher work, verifier calls, runtime, and peak memory, together with deployed parameter count.

\begin{table}[!htbp]
    \centering
    \caption{Reported training-account measurements and deployment size. Teacher samples are additional continuations generated specifically for teacher adaptation.}
    \label{tab:cost-accounting}
    \scriptsize
    \begin{tabular}{@{}lrrrrrrr@{}}
        \toprule
        Method & T. samples & Verifier & Tokens & GPU-h & Wall-h & Peak GB & Deploy B \\
        \midrule
        Vanilla OPSD & 0 & 400 & 220,476 & 0.0090 & 0.0034 & 53.837 & 1.7 \\
        SD-Zero & 0 & 13,456 & 7,548,980 & 0.3667 & 0.1400 & 60.701 & 1.7 \\
        H2SD & 0 & 3,200 & 1,914,827 & 0.0971 & 0.0371 & 61.734 & 1.7 \\
        PAST & 850 & 1,250 & 743,366 & 0.0392 & 0.0148 & 62.971 & 1.7 \\
        \bottomrule
    \end{tabular}
\end{table}

Within PAST, adaptive sampling reduces both samples and tokens relative to fixed-8 sampling, as reported in Table~\ref{tab:adaptive-sampling}. Full-vocabulary KL increases peak memory relative to sampled-token approximations; sequence chunking provides a lower-memory implementation while preserving close numerical agreement with the reference (Table~\ref{tab:full-vocab-implementation}).

\subsection{Reproducibility Boundary}

Each complete cycle checkpoint stores the student and teacher LoRA adapters; their optimizer, scheduler, and mixed-precision states; the next cycle identifier; logical-step counters; adaptive-sampling state; data cursor; and Python, PyTorch, CUDA, and rollout-sampler random-number-generator states. It also records the base checkpoint revision, tokenizer, chat template, verifier, and full training configuration. Resume begins at the next rollout phase. Partial-cycle rollouts, logits, gradients, and verifier buffers are discarded rather than promoted to a completed state.

Autoregressive student and teacher sampling use the corresponding named LoRA adapter in the rollout backend. Full-vocabulary distributions and all backward passes are computed in the training backend. Adapter version and cycle identifiers are checked when updated weights are synchronized. The CPU test backend implements the same phase and routing semantics without running large-model inference.
 
\end{document}